\def\year{2022}\relax
\documentclass[letterpaper]{article} 
\usepackage{aaai22}  
\usepackage{times}  
\usepackage{helvet}  
\usepackage{courier}  
\usepackage[hyphens]{url}  
\usepackage{graphicx} 
\usepackage{natbib}  
\usepackage{caption} 
\DeclareCaptionStyle{ruled}{labelfont=normalfont,labelsep=colon,strut=off} 
\usepackage{algorithm}
\usepackage{algorithmic}

\usepackage{newfloat}
\usepackage{listings}
\floatstyle{ruled}
\newfloat{listing}{tb}{lst}{}
\floatname{listing}{Listing}
\usepackage{array}
\usepackage{multirow}
\usepackage{cite}
\usepackage{amsmath}
\usepackage{amsfonts}
\usepackage{amssymb}

\title{Anchor DETR: Query Design for Transformer-Based Object Detection}
\author{
    Yingming Wang \qquad
    Xiangyu Zhang \qquad
    Tong Yang \qquad
    Jian Sun
}
\affiliations{
    MEGVII Technology  \\
    \{wangyingming, zhangxiangyu, yangtong, sunjian\}@megvii.com
}

\begin{document}

\maketitle

\begin{abstract}

In this paper, we propose a novel query design for the transformer-based object detection. 
In previous transformer-based detectors, the object queries are a set of learned embeddings.
However, each learned embedding does not have an explicit physical meaning and we cannot explain where it will focus on.
It is difficult to optimize as the prediction slot of each object query does not have a specific mode. 
In other words, each object query will not focus on a specific region.
To solved these problems, in our query design, object queries are based on anchor points, which are widely used in CNN-based detectors. 
So each object query focuses on the objects near the anchor point. 
Moreover, our query design can predict multiple objects at one position to solve the difficulty: ``one region, multiple objects''.
In addition, we design an attention variant, which can reduce the memory cost while achieving similar or better performance than the standard attention in DETR.
Thanks to the query design and the attention variant, the proposed detector that we called Anchor DETR, can achieve better performance and run faster than the DETR with 10$\times$ fewer training epochs.
For example, it achieves 44.2 AP with 19 FPS on the MSCOCO dataset when using the ResNet50-DC5 feature for training 50 epochs.
Extensive experiments on the MSCOCO benchmark prove the effectiveness of the proposed methods.
Code is available at \url{https://github.com/megvii-research/AnchorDETR}.

\end{abstract}


\begin{figure}[ht!]
	\centering
	\begin{tabular}{@{\hspace{0pt}}c}
		\includegraphics[width=0.9\columnwidth]{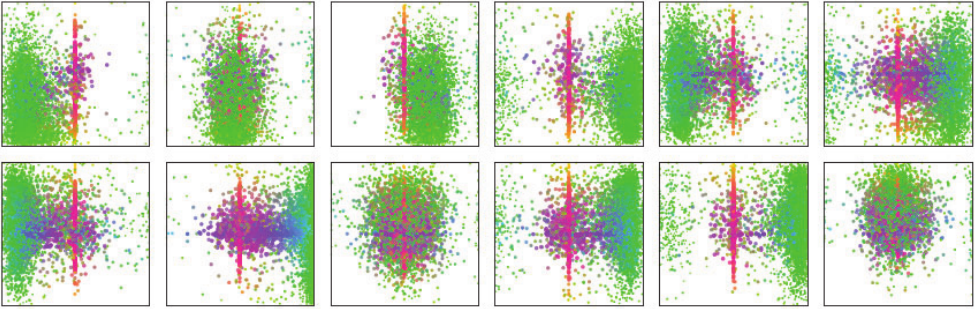}	     \vspace{0mm}  \\
	    (a) DETR\vspace{2mm}\\
		\includegraphics[width=0.9\columnwidth]{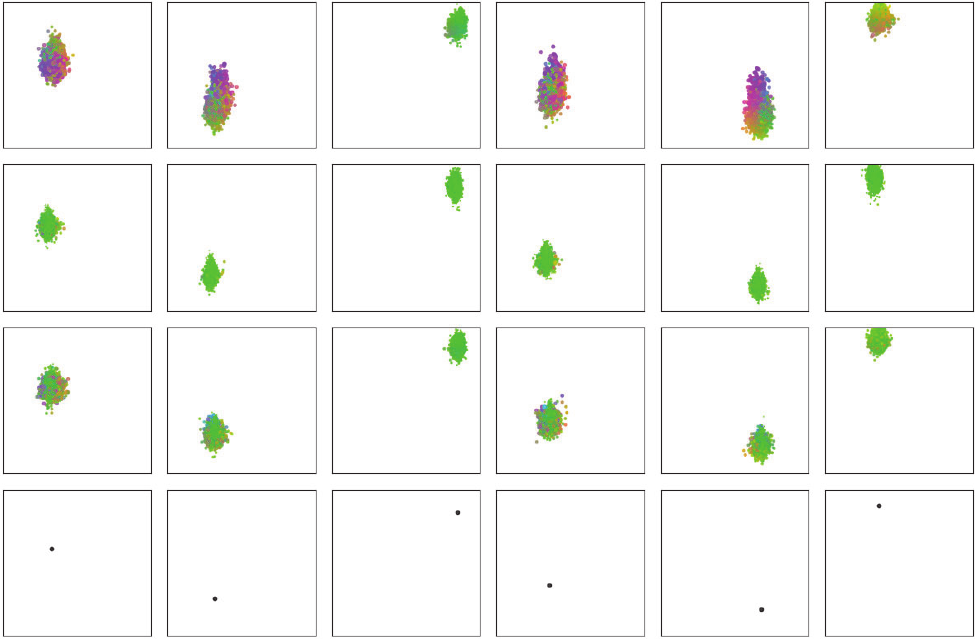}    \vspace{0mm}  \\
		(b) Ours\vspace{0mm}
	\end{tabular} 
	\caption{Visualization of the prediction slots. 
	Note that the sub-figure (a) comes from the figure in DETR~\cite{detr}.
	Each prediction slot includes all box predictions on the val set of a query. 
	Each of the colored points represents the normalized center position of a prediction. The points are color-coded so that green color corresponds to small boxes, red to large horizontal boxes, and blue to large vertical boxes. 
	The black points in the last row of sub-figure (b) indicate the anchor points.
	The prediction slots of ours are more related to a specific position than DETR.
    }
	\label{fig:query slots distribution}
\end{figure}

\section{Introduction}
\label{sec:introduction}

The object detection task is to predict a bounding box and a category for each object of interest in an image.
In the last decades, there are many great progresses in object detection based on the CNN~\cite{fastercnn,cascadercnn,yolo,retinanet,atss,detectors,yolof}.
Recently, Carion et al.~\cite{detr} propose the DETR which is a new paradigm of object detection based on the transformer.
It uses a set of learned object queries to reason about the relations of the objects and the global image context to output the final predictions set.
However, the learned object query is very hard to explain.
It does not have an explicit physical meaning and the corresponding prediction slots of each object query do not have a specific mode.
As shown in Figure~\ref{fig:query slots distribution}(a), the predictions of each object query in DETR are related to different areas and each object query will be in charge of a very large area.
This positional ambiguity, i.e., the object query does not focus on a specific region, makes it hard to optimize.

Reviewing the detectors based on CNN, the anchors are highly related to the position and contain interpretable physical meanings.
Inspired by this motivation, we propose a novel query design based on the anchor points, i.e., we encode the anchor points as the object queries.
The object queries are the encodings of anchor points' coordinates so that each object query has an explicit physical meaning.
However, this solution will encounter difficulty: there will be multiple objects appearing in one position.
In this situation, only one object query in this position cannot predict multiple objects, so that the object queries from other positions have to predict these objects collaboratively.
It will cause each object query to be in charge of a larger area.
Thus, we improve the object query design by adding multiple patterns to each anchor point so that each anchor point can predict multiple objects. 
As shown in Figure~\ref{fig:query slots distribution}(b), 
all the predictions of the three patterns of each object query are distributed around the corresponding anchor point.
In other words, it demonstrates that each object query only focuses on the objects near the corresponding anchor point.
So the proposed object query can be easy to explain.
As the object query has a specific mode and does not need to predict the object far away from the corresponding position, it can easier to optimize.

\begin{table}
\begin{center}
	\begin{tabular}{c|c@{\hspace{6pt}}c@{\hspace{6pt}}c@{\hspace{6pt}}c}
		\hline
	&Feature	&AP	&GFLOPs	&FPS	\\	\hline
DETR	&DC5	&43.3	&187	&12	\\	
SMCA	&multi-level	&43.7	&152	&10	\\	
Deformable DETR	&multi-level	&43.8	&173	&15	\\	
Conditional DETR	&DC5	&43.8	&195	&10	\\	
Ours	&DC5	&\textbf{44.2}	&172	&19	\\

		\hline
	\end{tabular}
\end{center}
\caption{Comparison with transformer detectors. The results are based on the ResNet-50 backbone and ``DC5'' means the dilated C5 feature. 
The DETR is trained with 500 epochs while the others are trained with 50 epochs.
We evaluate the FPS of the proposed detector by following the script in DETR and set the batch size to 1 for a fair comparison with others. Note that we follow the DETR to script the model to evaluate the speed.
}
\label{tab:detr}
\end{table}

Besides the query design, we also design an attention variant that we call Row-Column Decouple Attention (RCDA).
It decouples the 2D key feature into the 1D row feature and the 1D column feature, then conducts the row attention and column attention successively.
The RCDA can reduce the memory cost while achieving similar or better performance than the standard attention in DETR.
We believe it can be a good alternative to the standard attention in DETR.

As shown in Table~\ref{tab:detr}, thanks to the novel query design based on the anchor point and the attention variant, the proposed detector Anchor DETR can achieve better performance and run faster than the original DETR with even 10$\times$ fewer training epochs when using the same single-level feature.
Compared with other DETR-like detectors with 10$\times$ fewer training epochs, the proposed detector achieves the best performance among them.
The proposed detector can achieve 44.2 AP with 19 FPS when using a single ResNet50-DC5~\cite{resnet} feature for training 50 epochs.

The main contributions can be summarized as:
\begin{itemize}
	\item 
	We propose a novel query design based on anchor points for the transformer-based detectors.
	Moreover, we attach multiple patterns to each anchor point so that it can predict multiple objects for each position to deal with the difficulty: ``one region, multiple objects''.
	The proposed query based on anchor point is more explainable and easier to optimize than the learned embedding.
	Thanks to the effectiveness of the proposed query design, our detector can achieve better performance with 10$\times$ fewer training epochs than the DETR. 
	\item We design an attention variant that we called Row-Column Decoupled Attention. It can reduce the memory cost while achieving similar or better performance than the standard attention in DETR, which can be a good alternative to standard attention.
	\item Extensive experiments are conducted to prove the effectiveness of each component. 
\end{itemize}

\section{Relative Work}

\subsection{Anchors in Object Detection}
There are two type of anchors used in CNN-based object detectors, i.e. anchor boxes~\cite{fastercnn,retinanet} and anchor points~\cite{fcos,centernet}.
As the hand-craft anchor boxes need to be carefully tuned to achieve good performance, we may prefer to not using the anchor boxes.
We usually treat the anchor-free as the anchor boxes free so that the detectors using anchor point also be treated as anchor-free~\cite{fcos,centernet}.
DETR~\cite{detr} adopt neither the anchor boxes nor the anchor points.
It directly predicts the absolute position of each object in the image.
However, we find that introducing the anchor point into the object query can be better.

\subsection{Transformer Detector}
Vaswani et al.~\cite{attention} propose the transformer for sequence transduction at first.
Recently, Carion et al.~\cite{detr} propose the DETR which is based on the transformer for object detection.
The transformer detector will feed the information of value to the query base on the similarity of the query and key.
Zhu et al.~\cite{deformabledetr} propose the Deformable DETR with sampling deformable points of value to the query and using multiple level features to solve the slowly converge speed of the transformer detector.
Gao et al.~\cite{smca} add a Gaussian map in the original attention for each query.

Concurrent with us, the Conditional DETR~\cite{conditionaldetr} encodes the reference point as the query position embedding. 
But the motivations are different so that it only uses the reference point to generate position embedding as the conditional spatial embedding in the cross attention and the object queries are still the learned embeddings.
Besides, it does not involve the multiple predictions of one position and the attention variants.

\subsection{Efficient Attention}

The self-attention of the transformer has high complexity so that it cannot well deal with a very large number of queries and keys.
To solve this problem, many efficient attention modules have been proposed~\cite{linformer,efficientattention,attention,longformer,swin,luna}.
One method is to compute the key and value first which can lead to linear complexity of the number of the query or key.
The Efficient Attention~\cite{efficientattention} and the Linear Attention~\cite{katharopoulos2020transformers} follow this idea.
Another method is to restrict the attention region of the key for each query instead of the whole region.
The Restricted Self-Attention~\cite{attention}, Deformable Attention~\cite{deformabledetr}, Criss-Cross Attention~\cite{ccnet}, and LongFormer~\cite{longformer} follow this idea.
In this paper, we decouple the key feature to the row feature and the column feature by 1D global average pooling and then perform the row attention and the column attention successively.

\section{Method}

\subsection{Anchor Points}
\begin{figure}[htp]
	\centering
	\begin{tabular}{c@{\hspace{2mm}}c}
	\includegraphics[width=0.45\columnwidth]{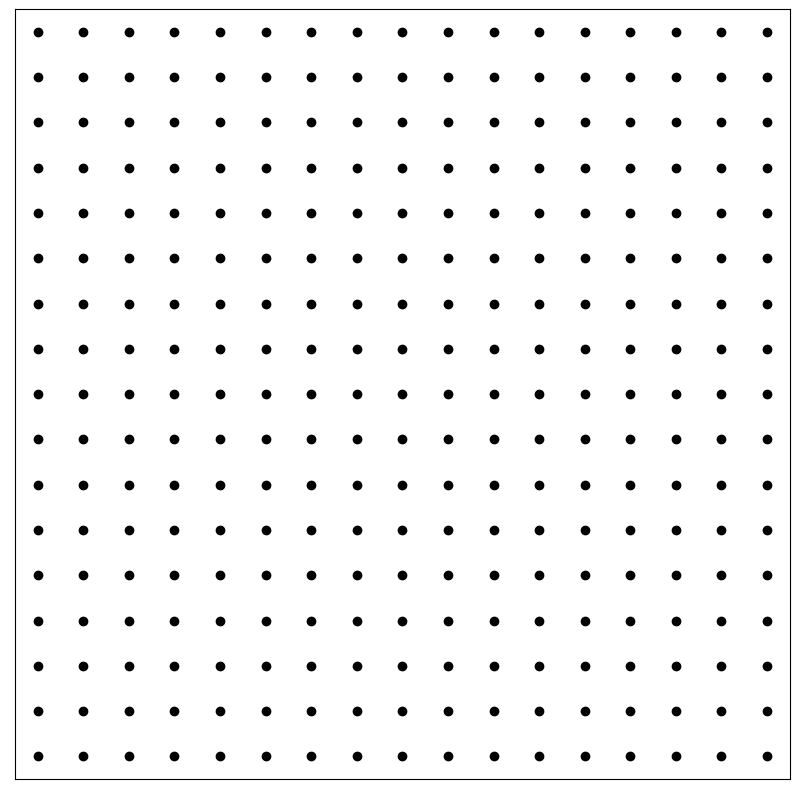}
	&\includegraphics[width=0.45\columnwidth]{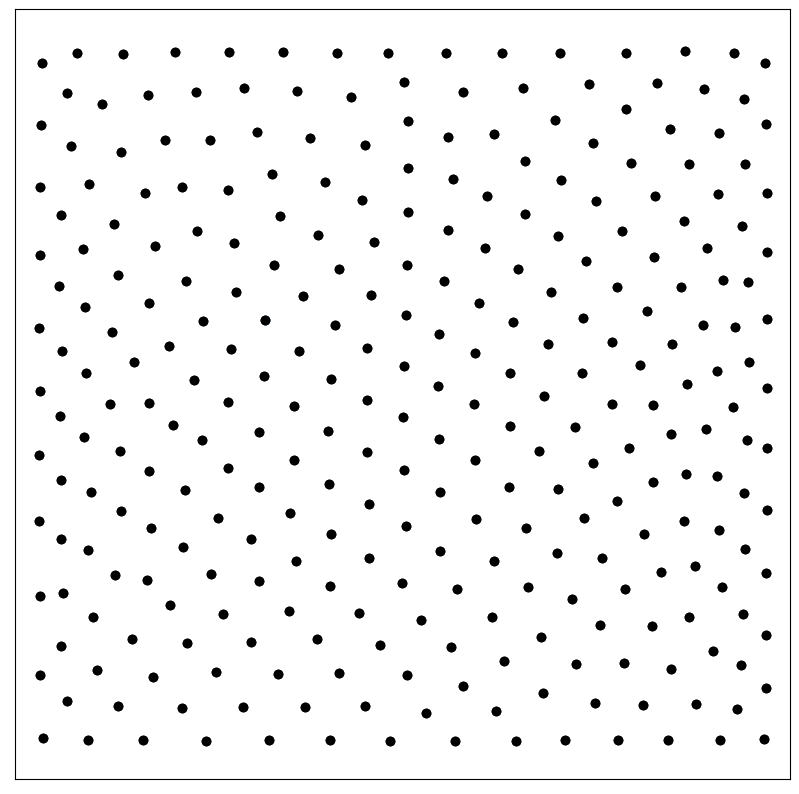}  \\
		(a) grid anchor points &(b) learned anchor points
	\end{tabular} 
	\caption{Visualization of the anchor points distribution. 
	Each point represents the normalized position of an anchor. 
	}
	\label{fig:query distribution}
\end{figure}

In the CNN-based detectors, anchor points always are the corresponding position of the feature maps.
But it can be more flexible in the transformer-based detector.
The anchor points can be the learned points, the uniform grid points, or other hand-craft anchor points.
We adopt two types of anchor points.
One is the grid anchor points and the other is the learned anchor points.
As shown in Figure~\ref{fig:query distribution}(a), the gird anchor points are fixed as the uniform grid point in the image.
The learned points are randomly initialized with uniform distribution from 0 to 1 and updated as the learned parameters.
With the anchor points, the center position $(\hat{cx},\hat{cy})$ of the predicted bounding box will be added to the corresponding anchor point as the final prediction like that in the Deformable DETR~\cite{deformabledetr}.

\subsection{Attention Formulation}
The attention of DETR-like transformer can be formulate as:
\begin{equation}
	\begin{split}
	\label{eq:att}
		\mathrm{Attention}(Q, K, V) = \mathrm{softmax}(\frac{QK^T}{\sqrt{d_k}})V,\\
		Q=Q_f+Q_p, K=K_f+K_p, V=V_f,
	\end{split}
\end{equation}
where the $d_k$ is the channel dimension, the subscript $f$ means the feature, and the subscript $p$ means the position embedding. The Q, K, V are the query, key, value respectively.
Note that the Q, K, V will pass through a linear layer respectively and that is omitted in Equation~(\ref{eq:att}) for clarity.

There are two attentions in the DETR decoder. 
One is self-attention and the other is cross-attention.
In the self-attention, the $K_f$ and the $V_f$ are the same as the $Q_f$ while the $K_p$ is the same as the $Q_p$.
The $Q_f\in \mathbb{R}^{N_q\times C}$ is the output of last decoder and the initial $Q_f^{init}\in \mathbb{R}^{N_q\times C}$ for the first decoder can be set to a constant vector or learned embedding.
For the query position embedding $Q_p\in \mathbb{R}^{N_q\times C}$, it uses a set of learned embedding in DETR:
\begin{equation}
	\label{eq:detr_qp}
	Q_p = \mathrm{Embedding}(N_q,C).
\end{equation}

In the cross-attention, the $Q_f\in \mathbb{R}^{N_q\times C}$ is generated from the output of the self-attention in front while the $K_f\in \mathbb{R}^{HW\times C}$ and $V_f\in \mathbb{R}^{HW\times C}$ are the output feature of the encoder. The $K_p \in \mathbb{R}^{HW\times C}$ is the position embedding of $K_f$. It is generated by the sine-cosine position encoding function~\cite{attention,detr} $g_{sin}$ based on the corresponding key feature position $Pos_k\in \mathbb{R}^{HW\times 2}$:
\begin{equation}
	\label{eq:detr_kp}
	K_p = g_{sin}(Pos_k).
\end{equation}
Note that the $H, W, C$ is the height, width, channel of the feature and the $N_q$ is the predefined number of the query.

\begin{figure*}[t]
	\centering
	\includegraphics[width=2\columnwidth]{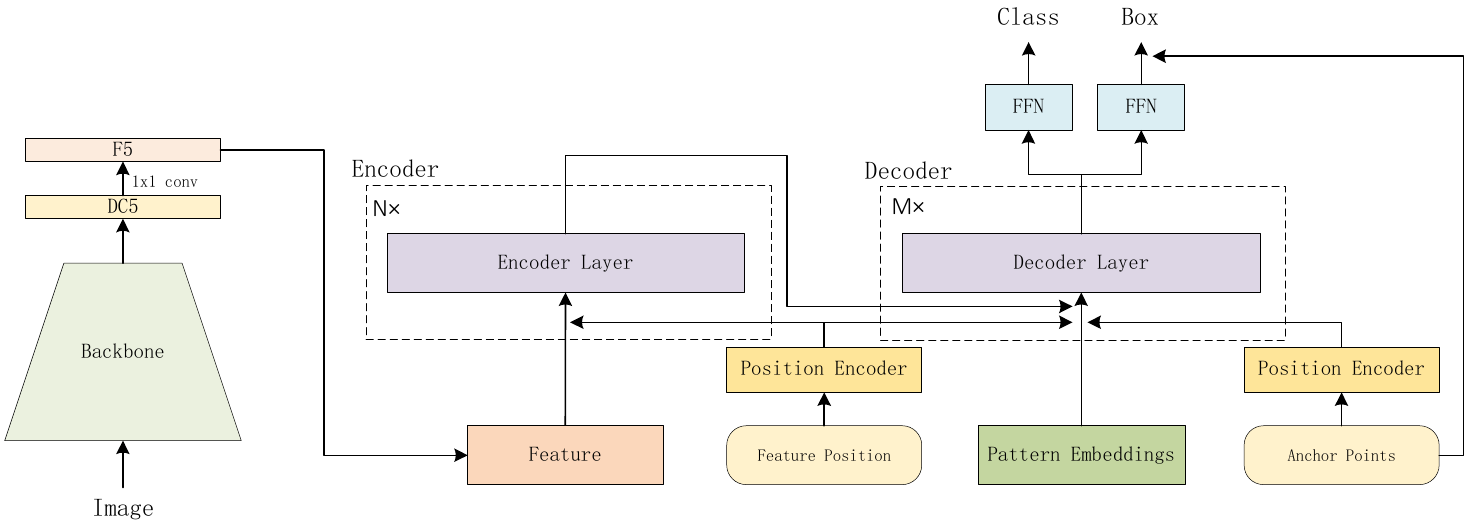} \\
	\caption{Pipeline of the proposed detector. Note that the encoder layer and decoder layer are in the same structure as DETR except that we replace the self-attention in the encoder layer and the cross-attention in the decoder layer with the proposed Row-Column Decouple Attention.}
	\label{fig:pipeline}
\end{figure*}

\subsection{Anchor Points to Object Query}

Usually, the $Q_p$ in decoder is regarded as the object query because it is responsible for distinguishing different objects.
The object query of learned embedding like Equation~(\ref{eq:detr_qp}) is hard to explain as discussed in the Introduction section.

In this paper, we propose to design the object query based on the anchor point $Pos_q$.
The $Pos_q\in \mathbb{R}^{N_{A}\times 2}$ represents $N_{A}$ points with their $(x, y)$ positions which range from 0 to 1.
Then the object query based on the anchor points can formulate as:
\begin{equation}
	\label{eq:pe}
	Q_p = Encode(Pos_q).
\end{equation}
It means that we encode the anchor points as the object queries.

So how to design the encoding function?
As the object query is designed as the query position embedding as shown in Equation~(\ref{eq:att}), the most natural way is to share the same position encoding function as key:
\begin{equation}
	\label{eq:cpe}
	Q_p = g(Pos_q),
	K_p = g(Pos_k),
\end{equation}
where $g$ is the position encoding function.
The position encoding function could be the $g_{sin}$ or other position encoding functions.
In this paper, instead of just using the heuristic $g_{sin}$, we prefer to use a small MLP network with two linear layers to adapt it additionally.

\subsection{Multiple Predictions for Each Anchor Point}
To deal with the situation that one position may have multiple objects, we further improve the object query to predict multiple objects for each anchor point instead of just one prediction.
Reviewing the initial query feature $Q_f^{init}\in \mathbb{R}^{N_q\times C}$, each of the $N_q$ object queries has one pattern $Q_f^i \in \mathbb{R}^{1\times C}$.  
Note the $i$ is the index of object queries.
To predict multiple objects for each anchor point, we can incorporate multiple patterns into each object query.
We use a small set pattern embedding $Q_f^i \in \mathbb{R}^{N_{p}\times C}$:
\begin{equation}
	Q_f^i = \mathrm{Embedding}(N_{p},C),
\end{equation}
to detect objects with different patterns at each position. The $N_{p}$ is the number of pattern which is very small, e.g. $N_{p}=3$.
For the property of translation invariance, the patterns are shared for all the object queries.
Thus we can get the initial $Q_f^{init} \in \mathbb{R}^{N_{p}N_{A} \times C}$ and $Q_p\in \mathbb{R}^{N_{p}N_{A} \times C}$ by sharing the 
$Q_f^{i} \in \mathbb{R}^{N_{p}\times C}$ to each of the 
$Q_p\in \mathbb{R}^{N_{A}\times C}$.
Here the $N_q$ is equal to $N_p \times N_A$.
Then we can define the proposed Pattern-Position query design as:
\begin{equation}
	Q=Q_f^{init} + Q_p.
\end{equation}
For the following decoder, the $Q_f$ is also generated from the output of last decoder like DETR.

Thanks to the proposed query design, the proposed detector has an interpretable query and achieves better performance than the original DETR with 10$\times$ fewer training epochs.

\begin{table*}[ht]
	\begin{center}
		\begin{tabular}{@{\hspace{2pt}}c|c@{\hspace{4pt}}c@{\hspace{4pt}}c@{\hspace{4pt}}c|cccccc}
			\hline

	&Anchor-Free	&NMS-Free	&RAM-Free	&Epochs	&$AP$	&$AP_{50}$	&$AP_{75}$	&$AP_s$	&$AP_m$	&$AP_l$	\\	\hline
RetinaNet	& 	& 	&\checkmark	&36	&38.7	&58.0	&41.5	&23.3	&42.3	&50.3	\\	
FCOS	&\checkmark	& 	&\checkmark	&36	&41.4	&60.1	&44.9	&25.6	&44.9	&53.1	\\	
POTO	&\checkmark	&\checkmark	&\checkmark	&36	&41.4	&59.5	&45.6	&26.1	&44.9	&52.0	\\	
Faster RCNN	& 	& 	& 	&36	&40.2	&61.0	&43.8	&24.2	&43.5	&52.0	\\	
Cascade RCNN	& 	& 	& 	&36	&44.3	&62.2	&48.0	&26.6	&47.7	&57.7	\\
Sparse RCNN	&\checkmark	&\checkmark	& 	&36	&\textbf{44.5}	&63.4	&48.2	&26.9	&47.2	&59.5	\\	
\hline
DETR-C5	&\checkmark	&\checkmark	&\checkmark	&500	&42.0	&62.4	&44.2	&20.5	&45.8	&61.1	\\	
DETR-DC5	&\checkmark	&\checkmark	&\checkmark	&500	&43.3	&63.1	&45.9	&22.5	&47.3	&61.1	\\	
SMCA	&\checkmark	&\checkmark	&\checkmark	&50	&43.7	&63.6	&47.2	&24.2	&47.0	&60.4	\\	
Deformable DETR	&\checkmark	&\checkmark	& 	&50	&43.8	&62.6	&47.7	&26.4	&47.1	&58.0	\\	
Ours-C5	&\checkmark	&\checkmark	&\checkmark	&50	&42.1	&63.1	&44.9	&22.3	&46.2	&60.0	\\	
Ours-DC5	&\checkmark	&\checkmark	&\checkmark	&50	&\textbf{44.2}	&64.7	&47.5	&24.7	&48.2	&60.6	\\	\hline
DETR-C5-R101	&\checkmark	&\checkmark	&\checkmark	&500	&43.5	&63.8	&46.4	&21.9	&48.0	&61.8	\\	
DETR-DC5-R101	&\checkmark	&\checkmark	&\checkmark	&500	&44.9	&64.7	&47.7	&23.7	&49.5	&62.3	\\	
Ours-C5-R101	&\checkmark	&\checkmark	&\checkmark	&50	&43.5	&64.3	&46.6	&23.2	&47.7	&61.4	\\	
Ours-DC5-R101	&\checkmark	&\checkmark	&\checkmark	&50	&\textbf{45.1}	&65.7	&48.8	&25.8	&49.4	&61.6	\\

			\hline
		\end{tabular}
	\end{center}
	\caption{Comparison with other detectors. The results are based on the ResNet-50 backbone if without specific and R101 means the ResNet-101 backbone. ``C5'', ``DC5'' indicate the detectors using a single C5 or DC5 feature respectively. 
	The other detectors without specific use multiple-level features.
		The RAM means the random access of memory which usually is not friendly to hardware in practice.
	}
	\label{tab:main result}
\end{table*}

\begin{table*}[ht]
	\begin{center}
		\begin{tabular}{ccc|cccccc}
			\hline

RCDA	&anchors	&patterns	&$AP$	&$AP_{50}$	&$AP_{75}$	&$AP_s$	&$AP_m$	&$AP_l$	\\	\hline
	& 	& 	&39.3	&59.4	&41.8	&20.7	&42.6	&55.0	\\	
	&\checkmark	&\checkmark	&\textbf{44.2}	&65.3	&47.2	&24.4	&47.8	&61.8	\\	
\checkmark	& 	& 	&40.3	&61.6	&42.5	&21.7	&44.1	&56.3	\\	
\checkmark	& 	&\checkmark	&40.3	&60.8	&43.0	&21.1	&44.2	&57.0	\\	
\checkmark	&\checkmark	& 	&42.6	&63.6	&45.5	&23.2	&46.4	&58.3	\\	
\checkmark	&\checkmark	&\checkmark	&\textbf{44.2}	&64.7	&47.5	&24.7	&48.2	&60.6	\\

			\hline
		\end{tabular}
	\end{center}
	\caption{Effectiveness of each component. The RCDA means to replace the standard attention with the RCDA. The ``anchors'' means using the anchor points to generate the object queries. The ``patterns'' means assigning multiple patterns to each object query so that each position has multiple predictions.
	}
	\label{tab:component}
\end{table*}

\begin{table*}[ht]
	\begin{center}
		\begin{tabular}{cc|cccccc}
			\hline
anchor points	&patterns	&$AP$	&$AP_{50}$	&$AP_{75}$	&$AP_s$	&$AP_m$	&$AP_l$	\\	\hline 
100	&1	&40.8	&61.5	&43.6	&21.0	&44.8	&57.5	\\	
100	&3	&43.4	&63.7	&46.5	&23.8	&47.1	&60.9	\\	
300	&1	&42.6	&63.6	&45.5	&23.2	&46.4	&58.3	\\	
300	&3	&\textbf{44.2}	&64.7	&47.5	&24.7	&48.2	&60.6	\\	
900	&1	&42.9	&63.8	&46.1	&24.4	&46.7	&58.9	\\	

			\hline
		\end{tabular}
	\end{center}
	\caption{Comparison for multiple predictions of one position. We show the performance with the different number of anchor points and patterns. Note that it degenerates to the single prediction of one anchor point when the number of the pattern is 1.
	}
	\label{tab:ppq}
\end{table*}

\begin{table}[ht]
	\begin{center}
		\begin{tabular}{c|cccccc}
			\hline
		&$AP$	&$AP_{50}$	&$AP_{75}$	&$AP_s$	&$AP_m$	&$AP_l$	\\	\hline 
		
grid	&44.1	&64.5	&47.6	&24.8	&48.0	&60.9	\\
learned	&\textbf{44.2}	&64.7	&47.5	&24.7	&48.2	&60.6	\\	

			\hline
		\end{tabular}
	\end{center}
	\caption{Comparison of different types of anchor points.
	The ``learned'' means that it uses 300 learned anchor points. The ``grid'' means that it uses the hand-craft grid anchor points with the number of $17 \times17$ which is near to 300.
	}
	\label{tab:anchor point}
\end{table}

\begin{table*}[ht]
	\begin{center}
		\begin{tabular}{c|c|cccccc|c}
			\hline
Attention	&Feature	&$AP$	&$AP_{50}$	&$AP_{75}$	&$AP_s$	&$AP_m$	&$AP_l$	&memory		\\	\hline
Luna~\cite{luna}	&\multirow{4}{*}{DC5}	&36.2	&58.7	&37.1	&15.4	&39.2	&54.7	&2.9G		\\	
Efficient-att~\cite{efficientattention}	& 	&34.1	&56.8	&34.4	&13.9	&36.0	&53.9	&2.2G		\\	
std-att	& 	&\textbf{44.2}	&65.3	&47.2	&24.4	&47.8	&61.8	&10.5G		\\	
RCDA	& 	&\textbf{44.2}	&64.7	&47.5	&24.7	&48.2	&60.6	&4.4G		\\	\hline
std-att	&\multirow{2}{*}{C5}	&42.2	&63.5	&44.9	&21.9	&45.9	&61.0	&2.7G		\\	
RCDA	& 	&42.1	&63.1	&44.9	&22.3	&46.2	&60.0	&2.3G		\\	

			\hline
		\end{tabular}
	\end{center}
	\caption{Comparison for different attention modules. The ``std-att'' means the standard attention in DETR. 
    The linear complexity attention modules can reduce memory significantly but the performance will drop.
	The proposed RCDA can achieve similar performance as the standard attention in DETR while saving the memory. }
	\label{tab:attention}
\end{table*}

\subsection{Row-Column Decoupled Attention}
\label{method:rcda}
The transformer will cost a lot of GPU memory which may limit it using the high-resolution feature or other extensions. 
The Deformable Transformer~\cite{deformabledetr} can reduce memory cost but it will lead to random access of memory which may not be friendly for the hardware.
There are also some attention modules~\cite{luna,efficientattention} with linear complexity and will not lead to random access of memory.
However, in our experiments, we find that these attention modules cannot well deal with the DETR-like detectors. It may be because the cross-attention in the DETR-like decoder is much difficult than the self-attention.

In this paper, we propose the Row-Column Decoupled Attention (RCDA) which can not only decrease the memory burdens but also achieve similar or better performance than the standard attention in DETR.
The main idea of the RCDA is to decouple the key feature $K_f\in \mathbb{R}^{H \times W\times C}$ to the row feature $K_{f,x}\in \mathbb{R}^{W\times C}$ and the column feature $K_{f,y}\in \mathbb{R}^{H\times C}$ by 1D global average pooling.
Then we perform the row attention and the column attention successively.
Without losing generality, we hypothesize $W\ge H$ and the RCDA can be formulated as:
\begin{equation}
	\begin{split}
		A_x= \mathrm{softmax}(\frac{Q_x K_{x}^T}{\sqrt{d_k}}), A_x \in \mathbb{R}^{N_q\times W},\\
		Z=\mathrm{weighted\_sumW}(A_x,V), Z \in \mathbb{R}^{N_q\times H \times C},\\
		A_y= \mathrm{softmax}(\frac{Q_y K_{y}^T}{\sqrt{d_k}}), A_y  \in \mathbb{R}^{N_q\times H},\\
		Out = \mathrm{weighted\_sumH}(A_y,Z), Out \in \mathbb{R}^{N_q\times C},
	\end{split}
\end{equation}
where 
\begin{gather}
	Q_x = Q_f + Q_{p,x},~~ Q_y  = Q_f + Q_{p,y},   \nonumber\\
	Q_{p,x} = g_{1D}(Pos_{q,x}),~~ Q_{p,y} = g_{1D}(Pos_{q,y}), \nonumber \\
	K_x= K_{f,x} + K_{p,x}, ~~K_y =  K_{f,y} + K_{p,y}, \\
	K_{p,x} = g_{1D}(Pos_{k,x}), ~~K_{p,y} = g_{1D}(Pos_{k,y}),  \nonumber\\
	V = V_f,~~V \in \mathbb{R}^{H \times W\times C}. \nonumber
\end{gather}
The $\mathrm{weighted\_sumW}$ and $\mathrm{weighted\_sumH}$ operations conduct the weighted sum along the width dimension and the height dimension respectively. The $Pos_{q,x} \in \mathbb{R}^{N_q\times 1}$ is the corresponding row position of $Q_f \in \mathbb{R}^{N_q\times C}$ and the $Pos_{q,y}\in \mathbb{R}^{N_q\times 1}$,  $Pos_{k,x}\in \mathbb{R}^{W\times 1}$,  $Pos_{k,y}\in \mathbb{R}^{H\times 1}$ are in the similar manner. The $g_{1D}$ is the 1D position encoding function which is similar to $g$ and it will encode a 1D coordinate to a vector with C channels.

Now we analyze why it can save the memory.
We hypothesize the head number $M$ of multi-head attention to 1 for clarity in the previous formulation without losing generality, but we should consider the head number $M$ for memory analysis.
The major memory cost for the attention in DETR is the attention weight map $A\in \mathbb{R}^{N_q\times H \times W \times M}$.
But the attention weight map of RCDA is $A_x \in \mathbb{R}^{N_q\times W \times M}$ and $A_y  \in \mathbb{R}^{N_q\times H \times M}$ whose memory cost is much smaller than the $A$.
However, the major memory cost in the RCDA is the temporary result 
$Z$. 
So we should compare the memory cost of $A\in \mathbb{R}^{N_q\times H \times W \times M}$ and $Z \in \mathbb{R}^{N_q\times H \times C}$.
The ratio for saving memory of the RCDA is:
\begin{equation}
	\begin{aligned}
		r&=(N_q\times H \times W \times M) / ( N_q\times H \times C) \\
		& = W \times M / C 
	\end{aligned}
\end{equation}
where the default setting is $M=8, C=256$. 
So the memory cost is roughly the same when the large side $W$ is equal to 32 which is a typical value for the C5 feature in object detection.
When using the high-resolution feature it can save the memory, e.g saving roughly 2x memory for the C4 or DC5 feature and saving 4x memory for the C3 feature.

\begin{figure*}[htp]
\centering
\begin{tabular}{c@{\hspace{2mm}}c@{\hspace{2mm}}c}
  \includegraphics[width=0.3\linewidth,height=0.3\linewidth]{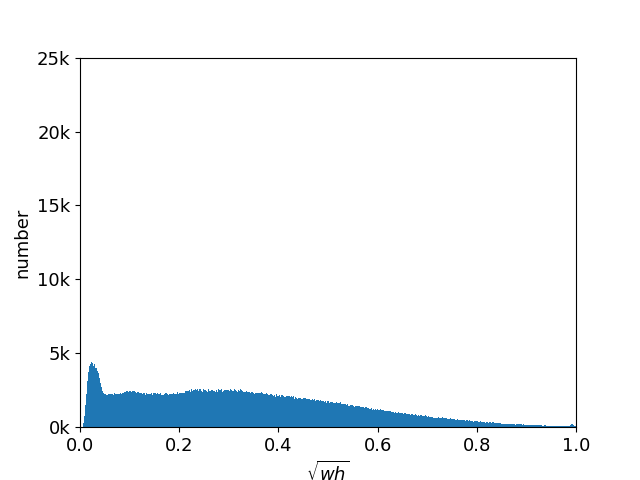}&
  \includegraphics[width=0.3\linewidth,height=0.3\linewidth]{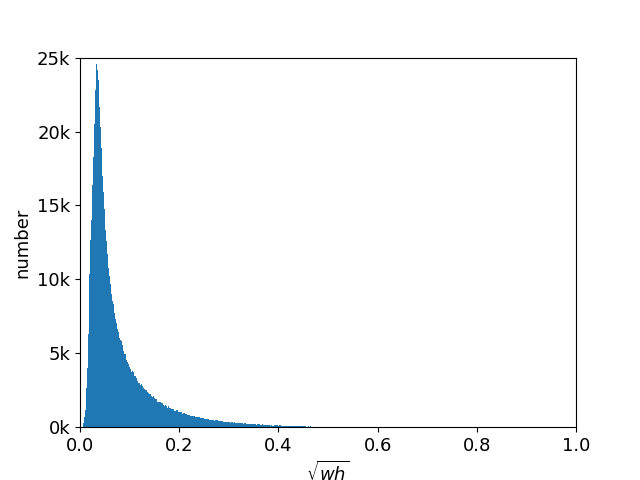}&
  \includegraphics[width=0.3\linewidth,height=0.3\linewidth]{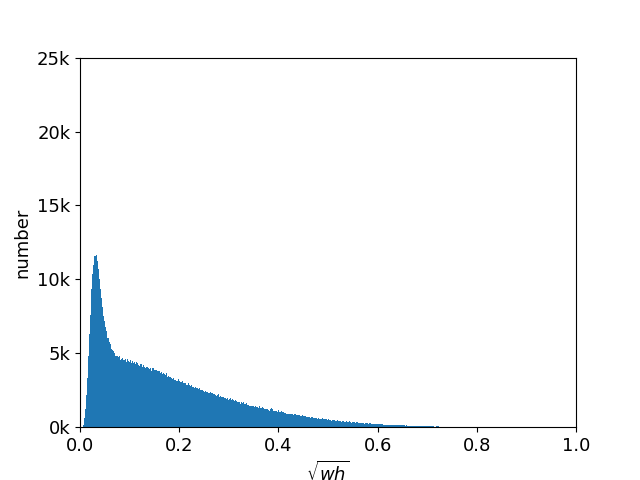}\\
~~(a) &~~(b) &~~(c)
\end{tabular}
  \caption{The histograms of each pattern. We show the histograms of predicted box sizes for different patterns. The abscissa represents the square root of the area of the predicted box with normalized width ``w'' and height ``h''. The large boxes usually appear in pattern (a), the pattern (b) focuses on the small objects, and the pattern (c) is in between.}
  \label{fig:pattern hist}
\end{figure*}

\section{Experiment}

\subsection{Implementation Details}
We conduct the experiments on the MS COCO~\cite{mscoco} benchmark.
All models are trained on the train2017 split and evaluated on the val2017.
The models are trained on 8 GPUs with 1 image per GPU.
We train our model on the training set for 50 epochs with the AdamW optimizer~\cite{adamw} setting the initial learning rate to $10^{-5}$ for the backbone and $10^{-4}$ for the others.
The learning rate will be decayed by a factor of 0.1 at the 40th epoch.
We set the weight decay to $10^{-4}$ and the dropout rate to 0.
The head for the attention is 8, the attention feature channel is 256 and the hidden dimension of the feed-forward network is 1024.
We choose the number of anchor points to 300 and the number of patterns to 3 by default. 
We use a set of learned points as anchor points by default.
The number of encoder layers and decoder layers is 6 like DETR. 
We use the focal loss~\cite{retinanet} as the classification loss following the Deformable DETR.

\subsection{Main Results}

As shown in Table~\ref{tab:main result}, we compare the proposed detector with RetinaNet~\cite{retinanet}, FCOS~\cite{fcos}, POTO~\cite{poto}, Faster RCNN~\cite{fastercnn}, 
Cascased RCNN~\cite{cascadercnn}, Sparse RCNN~\cite{sparsercnn}, 
DETR~\cite{detr}, SMCA~\cite{smca}, and Deformable DETR~\cite{deformabledetr}.
The ``C5'' and ``DC5'' mean that the detector uses a single C5 or dilated C5 feature while the other detectors use multiple-level features.
Using multiple-level features usually will have a better performance but cost more resources. 
Surprisingly, our detector with a single DC5 feature can achieve better performance than the Deformable DETR and SCMA which use multiple-level features.
The proposed detector can achieve better performance than DETR with 10$\times$ fewer training epochs.
It proves the proposed query design is very effective.

We also show the property of anchor-free, NMS-free, and RAM-free for each detector in Table~\ref{tab:main result}.
The anchor-free and NMS-free indicate the detector does not need the hand-craft anchor box and non-maximum suppression. 
The RAM-free means the detector will not involve any random access of memory which can be very friendly for hardware in practice.
The two-stage detectors always are not RAM-free because the region of interest (RoI) is stochastic to the hardware and the operation of the RoI-Align~\cite{maskrcnn}/RoI-Pooling~\cite{fastrcnn} will involve the random access of memory. 
The Deformable DETR is similar to the two-stage detector as the position of the sample point is stochastic to the hardware so that it is not RAM-free.
On the contrary, the proposed detector inherits the properties of anchor-free, NMS-free, and RAM-free from the DETR with an improvement of performance and fewer training epochs.

\subsection{Ablation Study}

\paragraph{Effectiveness of each component}
Table~\ref{tab:component} shows the effectiveness of each component that we proposed.
The proposed query design based on anchor points that can predict multiple objects for each position improves the performance from 39.3 AP to 44.2 AP.
The improvements are 4.9 AP that proves the query design with a specific focused area can be much easier to optimize. 
For the attention variant RCDA, it achieves a similar performance as the standard attention when adopting the proposed query design.
It proves that the RCDA will not degrade the performance with lower memory cost.
Besides, by applying the original query design as the DETR, the RCDA can achieve 1.0 improvements.
We think it is slightly easier to optimize as the attention map is smaller than the standard attention.
 This gain will vanish when adopting the proposed query design as the query design has the same effect that makes it easier to optimize. 
The object query based on anchor points with a single prediction has 2.3 AP improvements and the multiple predictions for each anchor point can have 1.6 further improvements. 
Applying the multiple predictions to the original object query in DETR will not improve the performance.
It is because the original object query is not highly related to the position thus it cannot get benefits from the ``one region, multiple predictions''.

\paragraph{Multiple Predictions for Each Anchor Point}
Table~\ref{tab:ppq} shows that multiple predictions for each anchor points play an essential role in query design.
For a single prediction (1 pattern), we find that 100 anchor points are not good enough and 900 anchor points get similar performance as 300 anchor points, thus we use 300 anchor points by default. 
The multiple predictions (3 patterns) can outperform the single prediction (1 pattern) by 2.6 and 1.6 AP for 100 and 300 anchor points respectively.
With the same number of the predictions, the multiple predictions for each anchor point (300 anchor points, 3 patterns) can outperform the single prediction (900 anchor points, 1 pattern) for 1.3 AP.
It indicates that the improvements of multiple predictions do not come from the increased number of predictions.
These results prove the effectiveness of multiple predictions for each anchor point.

\paragraph{Anchor Points Types}
We have tried two types of anchor points, i.e., the grid anchor points and the learned anchor points.
As shown in Figure~\ref{fig:query distribution}, the learned anchor points distribute uniformly in the image which is similar to the grid anchor points. 
We hypothesize that this is because the objects distribute everywhere in large COCO data set.
We also find the grid anchor points can achieve similar performance as the learned anchor points in Table~\ref{tab:ppq}.

\paragraph{Prediction Slots of Object Query}

As shown in Figure~\ref{fig:query slots distribution}(b), we can observe that the prediction slots of each object query in the proposed detector will focus on the objects near the corresponding anchor point.
As there are three patterns for the anchor points, we also show the histogram of each pattern in Figure~\ref{fig:pattern hist}.
We find that the patterns are related to the object size as the large boxes usually appear in pattern (a) while pattern (b) focuses on the small objects.
But the query patterns do not just depend on the size of the object because the small boxes can also appear in the pattern (a).
We think it is because there are more small objects, and multiple small objects are more likely to appear in one region.
So the all the patterns will be in charge of the small objects.

\paragraph{Row-Column Decoupled Attention}
As shown in Table~\ref{tab:attention}, we compare the Row-Column Decoupled Attention with the standard attention in DETR and some efficient attention modules with linear complexity.
The attention modules with linear complexity can significantly reduce the training memory compared to the standard attention module.
However, their performance is decreased by about 10 AP compared to the standard attention module.
It seems these efficient attentions are not suitable for the DETR-like detector.
On the contrary, the Row-Column Decoupled Attention achieves a similar performance.
As previously discussed, the Row-Column Decoupled Attention can significantly reduce memory when using the high-resolution feature and get roughly the same memory cost when using the C5 feature.
For example, RCDA reduces the memory from 10.5G to 4.4G and gets the same performance as the standard attention when using the DC5 feature.
In conclusion, the Row-Column Decoupled Attention can be efficient in memory while preserving competitive performance so that it can be a good alternative to the standard attention in DETR.

\section{Conclusion}
In this paper, we propose a detector based on the transformer.
We propose a novel query design based on the anchor point that has explicit physical meaning. 
The corresponding prediction slots can have a specific mode, i.e. the predictions are near the anchor point, so that it is easier to optimize.
Moreover, we incorporate multiple patterns to each anchor point to solve the difficulty: ``one region, multiple objects''.
We also propose an attention variant named the Row-Column Decoupled Attention.
The Row-Column Decouple Attention can reduce the memory cost while achieving similar or better performance than the standard attention in DETR.
The proposed detector can achieve better performance and run faster with 10$\times$ fewer training epochs than the DETR.
The proposed detector is an end-to-end detector with the property of anchor-free, NMS-free, and RAM-free.
We hope the proposed detector can be useful in practice and be a strong simple baseline for research.

\section{Acknowledgments}
This paper is supported by the National Key R\&D Plan of the Ministry of Science and Technology (Project No. 2020AAA0104400) and Beijing Academy of Artificial Intelligence (BAAI).

\bibliography{aaai22}

\end{document}